# Image Colour Segmentation by Genetic Algorithms

Vitorino Ramos, Fernando Muge

*CVRM - Centro de Geosistemas, Instituto Superior Técnico,*
*Av. Rovisco Pais, Lisboa, PORTUGAL*
*{vitorino.ramos,muge@alfa.ist.utl.pt}*

*Abstract*: Segmentation of a colour image composed of different kinds of texture regions can be a hard problem, namely to compute for an exact texture fields and a decision of the optimum number of segmentation areas in an image when it contains similar and/or unstationary texture fields. In this work, a method is described for evolving adaptive procedures for these problems. In many real world applications data clustering constitutes a fundamental issue whenever behavioural or feature domains can be mapped into topological domains. We formulate the segmentation problem upon such images as an optimisation problem and adopt evolutionary strategy of Genetic Algorithms for the clustering of small regions in colour feature space. The present approach uses k-Means unsupervised clustering methods into Genetic Algorithms, namely for guiding this last Evolutionary Algorithm in his search for finding the optimal or sub-optimal data partition, task that as we know, requires a non-trivial search because of its intrinsic NP-complete nature. To solve this task, the appropriate genetic coding is also discussed, since this is a key aspect in the implementation. Our purpose is to demonstrate the efficiency of Genetic Algorithms to automatic and unsupervised texture segmentation. Some examples in Colour Map, Ornamental Stones and in Human Skin Mark segmentation are presented and overall results discussed.

## 1. INTRODUCTION

Image segmentation is a low-level image processing task that aims at partitioning an image into homogeneous regions [11]. How region homogeneity is defined depends on the application. A great number of segmentation methods are available in the literature to segment images according to various criteria such as for example grey level, colour, or texture. This task is hard and as we know very important, since the output of an image segmentation algorithm can be fed as input to higher-level processing tasks, such as model-based object recognition systems. Recently, researchers have investigated the application of genetic algorithms (GA, [13,8,15]) into the image segmentation problem. Perhaps the most extensive and detailed work on GAs within image segmentation is that of *Bhanu* and *Lee* [3]. Many general pattern recognition applications of this particular paradigm can also be found in [16]. One reason (among others) for using this kind of approach is mainly related with the GA ability to deal with large, complex search spaces in situations where only minimum knowledge is available about the objective function. For example, most existing image segmentation algorithms have many parameters that need to be adjusted. The corresponding search space is in many situations, quite large and there are complex interactions among parameters, namely if we are seeking to solve colour image segmentation problems. For instance, this led *Bhanu et al*. [3] to adopt a GA to determine the parameter set that optimise the output of an existing segmentation algorithm under various conditions of image acquisition. That was the case for the optimisation of the *Phoenix* segmentation algorithm [22], by genetic algorithms, implementation described also by *Bhanu* [4]. Another situation wherein GAs may be useful tools is illustrated by the work of *Yoshimura* and *Oe* [23]. In their work, the two authors formulated the segmentation problem upon textured images as an optimisation problem, and adopt GAs for the clustering of small regions in a feature space, using also *Kohonen's* self-organising maps (SOM). They divided the original image into many small rectangular regions and extracted texture features from the data in each small region by using the two-dimensional autoregressive model (2D-AR), fractal dimension, mean and variance. In other example, *Bhandarkar et al*. [2] defined a multi-term cost function, which is minimised using a GA-evolved edge configuration. The idea was to solve medical image problems, namely edge-detection. In their approach to image segmentation, edge detection is cast as the problem of minimising an objective cost function over the space of all possible edge configurations and a population of edge images is evolved using specialised operators. Results comparable with those obtained using simulated annealing are reported. Fuzzy GA fitness functions were also considered by *Chun* and *Yang* [7], mapping a region-based segmentation onto the binary string representing an individual, and evolving a population of possible segmentations.

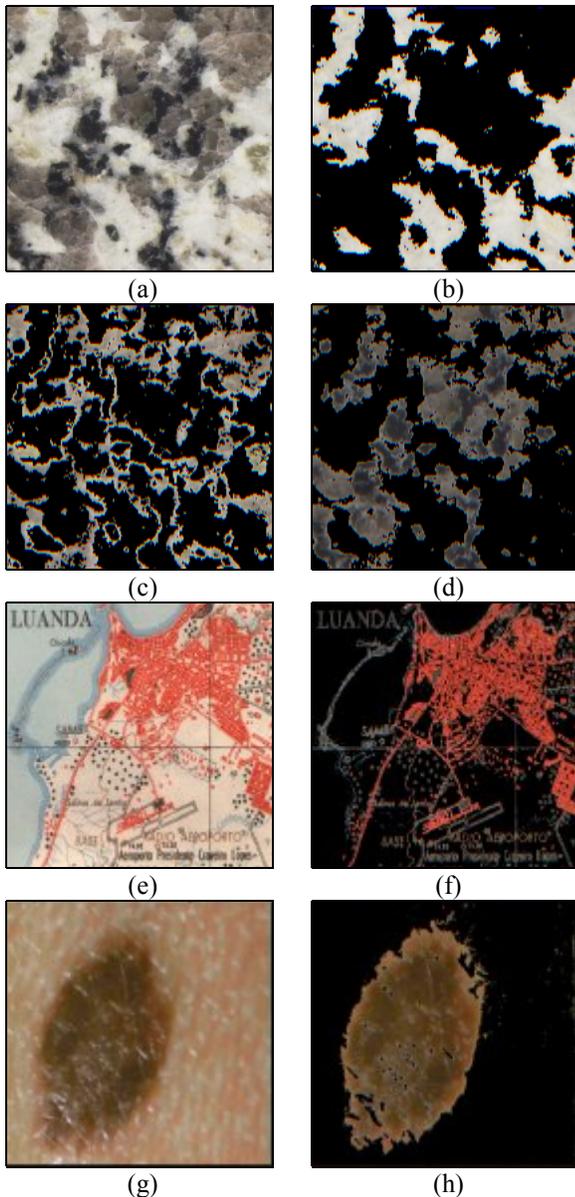

Figure 1 - (a) Original Colour Portuguese *Branco Coral* granite, and the respective 3 colour clusters in (b-whiter, c-greyer, and d-darker pixels). (e) Original Luanda (Angola) Colour Map and (e) the respective $5^{th}$ colour cluster segmenting roads (in red), names (in black and brown), and parks (in dark green). (g) Original human skin cancer mark (thanks to Prof. *Caldas Pinto* - IST) mainly detected by perimeter and colour intensity evolution on time, and (h) the first colour cluster pointed by the GA (the second colour is dual to (h) / even noisy flash lights in image acquisition were properly segmented).

Other implementations include the search of optimal descriptors to represent 3D structures [10], or the optimisation of parameters in GA hybrid systems [17] - in this last case, for finding the appropriate parameters of recurrent neural networks to segment echocardiographic images. GA applications within elastic-contour models are also possible to find. *Cagnoni et al*. [6] develop a GA based on a small set of manually-traced contours of the structure of interest (anatomical structures in 3D medical data sets). As putted by the authors, the method combines the good trade-off between simplicity and versatility offered by polynomial filters with the regularisation properties that characterise elastic-contour models. Another very interesting work, is that one of *Andrey* [1]. The image to be segmented is considered as an artificial environment wherein regions with different characteristics according to the segmentation criterion are as many ecological niches. A GA is then used to evolve a population of chromosomes that are distributed all over this environment. Each chromosome belongs to one out of a number of distinct species. The GA-driven evolution leads distinct species to spread over different niches. Consequently, the distribution of the various species at the end of the run unravels the location of the homogeneous regions on the original image. Because the segmentation progressively emerges as a by-product of a relaxation process [9] mainly driven by selection, the method has been called selectionist relaxation. In model designing terms, this last approach is indeed very close to that one presented by *Ramos* in [20], using artificial ant colonies and mathematical morphology extracted features. Approaches based on *Koza*'s genetic programming paradigm (GP, [14]), i.e., genetic algorithms used for finding appropriate algorithm structures and strategies, were also applied in image segmentation. *Poli*'s GP work [18], is perhaps one of the most interesting to follow, due to is simplicity. Finally, a fairly comprehensive review of other GA approaches in image processing is available in [5] - references include, animation, classification, feature extraction, filtering, image analysis, image processing, pattern recognition and naturally, image segmentation.

## 2. GENETIC CLUSTERING IN IMAGE SEGMENTATION

As putted by *Andrey*, whether the GA is used to search in the parameter space of an existing segmentation algorithm [4], or in the space of candidate segmentations [2], an objective fitness function, assigning a score to each segmentation, has to be specified in both cases. However, evaluating a segmentation result is itself a difficult task. To date, no standard evaluation method prevails [25], and different measures may yield distinct rankings [24] (as an aside note, the present authors are nowadays developing image noise measures by mathematical morphology [21], allowing for instance, his use in image filtering design by GAs). One possible criterion is to think of homogeneous regions as the result of any appropriate and optimised clustering process, within the image feature space. Applications of GAs in clustering and grouping problems are intensively described in [12]. In the present approach, grey level intensities of RGB image channels are considered as feature vectors, and the *k*-mean clustering model (*J.MacQueen*, 1967) is then applied as a quantitative criterion (or GA objective fitness function), for guiding the evolutionary algorithm in his appropriate search. Since the *k*-mean clustering model allows to minimise the internal feature variance of each colour cluster (or the maximisation of the variances between different colour clusters [19]), natural and homogeneous clusters can emerge if the GA is properly coded. In other words, the image segmentation problem is simply reformulated as an unsupervised clustering problem, and genetic algorithms are then used for finding the most appropriate and natural clusters. Since the clustering task can not be successfully applied within the image 2D space itself (e.g., similar pixels can be very far apart) the problem is coded within another space - that one of their colour features - 3D (grey level intensities, for the three channels). By this reformulation, one can in fact guarantee that similar pixels will belong to the same colour cluster. Preliminary efforts for this overall approach were designed on a previous attempt by *Ramos* in 1997 [19].

## 3. *K*-MEANS CLUSTERING MODEL

*K*-Means clustering models were introduced in 1967 by *J. MacQueen*, and they are considered as an unsupervised classification technique. Once the method uses a minimum distance criteria, many authors consider this approach in many ways similar to the k-nearest neighbour rule method (there is also some similarities with the *Kohonen* LVQ method). In *MacQueen* terms however, *k* stands for the number of clusters searched by the model (and given as an input). All the strategy undergoes the minimisation of the expression (1). If one admits the partition of a *p*-dimensional space into *c* clusters (*c* colour classes), where *n* samples exists (*n* points characterised by *p* features), and being $C_i$ each cluster *i* centre ($i = 1,…,c$), and $X_j$ representative of each sample *j* co-ordinates ($j = 1,…,n$), where $u_{ij}$ represents the hypothetical belonging of sample *j* into cluster *i* (i.e., $u_{ij} = 1$ if *j* belongs to cluster *i*; $u_{ij} = 0$ if *j* belongs to any other cluster different from *i*). Naturally for the present case, the idea is to compute the colour cube partition minimising *J* by using genetic algorithms. We then have *J* as in Eq.1:

$$u_{ij} = 0,1$$
$$u_{ij} \in U_{c \times n}$$
$$\sum_{i=1}^{c} u_{ij} = 1$$
$$\sum_{i=1}^{c} \sum_{j=1}^{n} u_{ij} = n \qquad (1)$$
$$C_i = \frac{1}{n_i} \sum_{j=1}^{n} X_j$$
$$\min J = \sum_{i=1}^{c} \sum_{j=1}^{n} u_{ij} \|X_j - C_i\|^2$$

## 4. GENETIC IMPLEMENTATION

The above minimisation is then based on the different belonging combinations, of all points in the feature space. Naturally that, such task will be simply if the number of colours in one image to segment is low; however for high number of points in this 3D colour space (i.e., the different number of colours) this minimisation is hard to compute, since the combinatorial search space becomes very large. However, the partition of this 3D histogram into different clusters, must take in account the value of each point (that is, his frequency for a given RGB point). By this, the minimisation described in section 3, suffers a little modification (the method becomes weighted by this frequency *f*, since the number of colours of any RGB point are an important information in the overall process). The above *J* expression then becomes (Eq.2):

$$\min J = \sum_{i=1}^{c}\sum_{j=1}^{n} u_{ij} f_j \left\| X_j - C_i \right\|^2 \qquad (2)$$

Another important issue in the GA implementation is the problem genetic coding. In order to do it appropriately, each chromosome codes the binary values of $u_{ij}$. However, to improve the GA search time and since the number of different colours in one colour image can be high, each 3D feature colour was submitted to a pre-partition. By this pre-procedure, the combinatorial search space is reduced, as also as the number of bits in each GA chromosome. That is, each 3D colour cube (with side 256 - 8 bit images) that could represent up to $256^3$ colours, was reduced to a maximum of 512 points (i.e., 512 small cubes with side 32). In other words, all RGB points that fall into a small cube are agglomerated, being the new point represented by the centre of this small cube, and his frequency being equal to the sum of all frequencies of those points.

## 5. RESULTS, CONCLUSIONS AND FUTURE WORK

The above strategy was applied in 3 different cases. A segmentation of ornamental stones (29349 colours), of human skin marks (303 colours) and in colour maps (214385 colours-see figure 1). Since the granite *Branco Coral* has 3 different types of minerals the aim was to search for 3 colour clusters. For the same reason, the skin mark segmentation was implemented with 2 clusters, and finally the colour maps into 6 clusters (6 prominent colours). Number of GA individuals were always 50, generations run = 10000, crossover mutation rate = 0.95, and mutation rate = 0.85. The respective computer time (PENTIUM 166 MHz / 32 Mb Ram) were 14.96, 12.76 and 37.02 minutes. Finally, each string length (a parameter that depends only in the number of different colours present) was respectively 124, 64 and 468 bits long. Overall results points to highly satisfactory results namely for the segmentation of ornamental stones and for the case of human skin mark segmentation. There is however some problems with the colour map examples. The main reason is that the problem is by itself difficult (with large combinatorial search spaces), and that the pre-partition tends to reduce the discriminatory power of the overall strategy. This is mainly observed within pixels that form bounds of any important colour object. Image acquisition with low resolutions interpolates somehow their grey level intensities into intermediate values (between inner and outer bounds), and the result (with pre-partition) is significantly altered, since similar pixels can belong to different small cubes (naturally with low probability). However if the number of this kind of pixels is high, the strategy tends to create himself another cluster. On the other hand, and by observing the GA performance (*J* value) in each generation, we can conclude that similar results can be achieved with half of the generations run, since after 5000 generations *J* values are increasing very slow. Future work includes three main lines. First, to study the cluster relations (clouds of points) for each segmentation problem. This can bring useful information into the GA approach, and simply neighbourhood relations can be computed by using mathematical morphology on the 3D-colour cube. Second, more relevant segmentation measures must be studied. The present authors are making nowadays preliminary attempts, even if they are related with image noise [21]. Finally, significant improvements on the automatic design can be achieved by using ISODATA models - number of clusters can be automatically chosen by the hybrid search model.